\documentclass[journal]{IEEEtran}
\ifCLASSINFOpdf
  \usepackage[pdftex]{graphicx}
\else
  \usepackage[dvips]{graphicx}
\fi
\usepackage{amsmath}
\usepackage{ulem}
\usepackage{color,soul}
\usepackage{bm}
\usepackage{booktabs}

\graphicspath{{./figures}}
\DeclareGraphicsExtensions{.pdf,.jpeg,.png, .eps}
\usepackage{epstopdf}
\usepackage{subfig}
\usepackage{booktabs} 
\usepackage{tabularx} 
\usepackage{siunitx} 
\usepackage{graphicx} 
\usepackage{tabularray}
\usepackage{enumitem}

\usepackage[hidelinks]{hyperref} 
\usepackage{amssymb}

\begin{document}
%
\title{How Certain are Uncertainty Estimates? Three Novel Earth Observation Datasets for Benchmarking Uncertainty Quantification in Machine Learning}
%
%
%

\author{Yuanyuan~Wang,
        Qian~Song,
        Dawood Wasif,
        Muhammad Shahzad,
        Christoph Koller,
        Jonathan Bamber,
        and~Xiao~Xiang~Zhu
\thanks{This work is supported by the German Federal Ministry of Education and Research in the framework of the international future AI lab "AI4EO -- Artificial Intelligence for Earth Observation: Reasoning, Uncertainties, Ethics and Beyond" (grant number: 01DD20001). The authors gratefully acknowledge the Gauss Centre for Supercomputing e.V. (www.gauss-centre.eu) for funding this project by providing computing time on the GCS Supercomputer SuperMUC-NG at Leibniz Supercomputing Centre (www.lrz.de).}
\thanks{Corresponding author: Yuanyuan Wang, Xiao Xiang Zhu.}
\thanks{Y. Wang, Q. Song, D. Wasif, M. Shahzad, C. Koller, J. Bamber, and X. X. Zhu are with the Chair Data Science in Earth Observation, Technical University of Munich, Munich, Germany. }
\thanks{M. Shahzad is also with the National University of Sciences \& Technology Pakistan. C. Koller is also with the Remote Sensing Technology Institute, German Aerospace Center. J. Bamber is also with the University of Bristol. X. Zhu is also with the Munich Center for Machine Learning, Munich, Germany}}

%
%

\markboth{Submitted to IEEE Geoscience and Remote Sensing Magazine,~ 2024}%
{Shell \MakeLowercase{\textit{et al.}}: Bare Demo of IEEEtran.cls for IEEE Journals}
%



\maketitle

\begin{abstract}
Uncertainty quantification (UQ) is essential for assessing the reliability of Earth observation (EO) products. However, the extensive use of machine learning models in EO introduces an additional layer of complexity, as those models themselves are inherently uncertain. While various UQ methods do exist for machine learning models, their performance on EO datasets remains largely unevaluated. A key challenge in the community is the absence of the ground truth for uncertainty, i.e. how certain the uncertainty estimates are, apart from the labels for the image/signal.
This article fills this gap by introducing three benchmark datasets specifically designed for UQ in EO machine learning models. These datasets address three common problem types in EO: regression, image segmentation, and scene classification. They enable a transparent comparison of different UQ methods for EO machine learning models. We describe the creation and characteristics of each dataset, including data sources, preprocessing steps, and label generation, with a particular focus on calculating the reference uncertainty. We also showcase baseline performance of several machine learning models on each dataset, highlighting the utility of these benchmarks for model development and comparison. Overall, this article offers a valuable resource for researchers and practitioners working in artificial intelligence for EO, promoting a more accurate and reliable quality measure of the outputs of  machine learning models. The dataset and code are accessible via \url{https://gitlab.lrz.de/ai4eo/WG_Uncertainty}.

\end{abstract}

\begin{IEEEkeywords}
uncertainty quantification, datasets, earth observation, benchmark
\end{IEEEkeywords}

%
\IEEEpeerreviewmaketitle

\section{Introduction}
%
%
%
%
\subsection{Motivation}
Researchers face a wide range of challenges in Earth observation (EO) downstream tasks, including tasks like regression (e.g., predicting crop yield), image segmentation (e.g., delineating building boundaries), and image classification (e.g., identifying land use land cover types). Due to the inherent complexity of many problems, traditional physical models often do not exist. This has led the scientific community to increasingly rely on machine learning models to extract meaningful insights from the vast amount of EO data. Deep learning techniques are at the forefront of development in data-intensive science in EO \cite{zhu_deep_2017}. Deep learning, especially convolutional neural networks (CNNs) and recurrent neural networks (RNNs), has proven to be very successful in various EO tasks, such as image recognition, object detection, semantic segmntation, action recognition, image captioning, and many more \cite{zhu_deep_2017,zhang_deep_2016}. Recently, the field has witnessed the rise of transformers and variants like vision transformers \cite{vaswani_attention_2017,dosovitskiy_image_2021}, a deep learning architecture inspired by how language models understand context. Unlike CNNs and RNNs, transformers don't rely on a specific order for processing data. 
Their potential extends beyond analyzing static snapshots of our planet, but also into Earth system science. 
Here, transformers emerges as a promising tool. One particularly noteworthy application of transformers in Earth system science is EarthFormer \cite{gao_earthformer_2022}. This model tackles a specific challenge: Earth system forecasting, which traditionally relies on complex physical models. EarthFormer leverages a  transformers ability to capture long-range dependencies to analyze spatio-temporal EO data, potentially offering a data-driven alternative for forecasting tasks.

EO data are increasingly being relied upon for a diverse range of problems in Earth system science including numerical weather prediction, climate services and in testing and tuning Earth System Model (ESM) projections \cite{vance_big_2024}. For example, the World Meteorological Organization has established, through its Global Climate Observing System, 55 Essential Climate Variables (ECVs). These are defined as physical, chemical or biological variables that are critical for defining Earth's climate and state. Around 60\% of these ECVs can be observed using EO data. In addition, EO data are providing unique insights into Earth surface processes and climate dynamics that are being used to test, tune and parameterize ESMs \cite{schneider_earth_2017,reichstein_deep_2019}. At the same time, we are experiencing an unprecedented growth in EO capabilities and the number of satellites in orbit, largely driven by the availability of more affordable commercial launcher platforms and satellites \cite{vance_big_2024}. This increase in volume and complexity of EO data necessitates new and efficient methods for information extraction, compression and dimensional reduction. Machine learning methods have been used for some time to achieve this \cite{salcedo-sanz_machine_2020} but with little attention placed on the robustness and uncertainties in the models used and the resulting outputs.

\begin{figure}[h]
\centering
\includegraphics[width=0.5\textwidth]{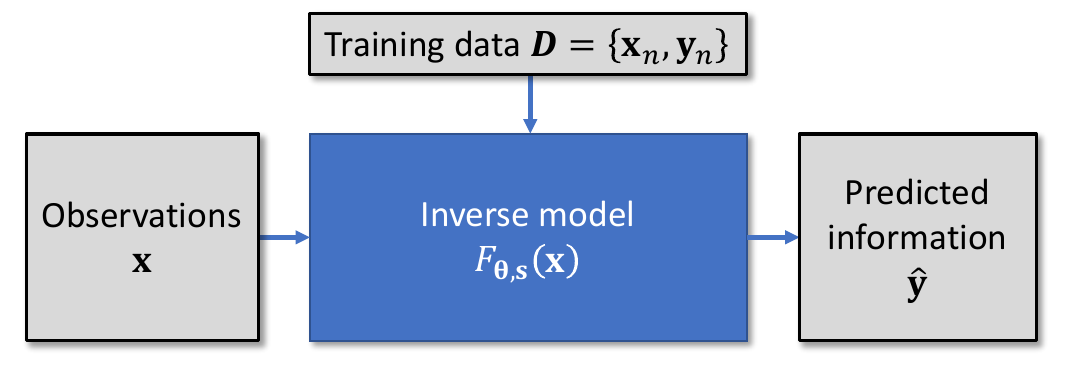}

\caption{Error sources of predictive uncertainty in EO with a machine learning model: 1. the noise in the observations $\mathbf{x}$, 2. the structural uncertainty in the model $\textit{F}$, e.g. its network architecture, and 3. the noise in the training data $\textbf{\textit{D}}$.}
\label{fig:model}
\end{figure}

As EO products are employed in various important activities, uncertainty quantification (UQ) has always been a crucial topic in EO. This is especially true when machine learning models are employed in information retrieval. EO involves collecting data from diverse sources, including satellites, airborne platforms, and ground-based sensors, that are subject to various sources of uncertainty. These uncertainties can arise from multiple factors such as varying sensor types, sensor calibration, spatial and temporal changes (e.g., changing illumination, weather conditions and different seasons), atmospheric interference, geometric distortions, and limitations in the measurement process. Those data are the inputs to the training and inference of machine learning models. The uncertainty originating from the input data is known as the data uncertainty, or \textit{aleatoric uncertainty} in machine learning \cite{gawlikowski2022survey}. Apart from that, machine learning models are inherently uncertain. The model architectures themselves are not guaranteed to capture the real physics or processes. This is also known as model uncertainty, or \textit{epistemic uncertainty}.  Summarized in Fig. \ref{fig:model}, the predictive uncertainty in  $\hat{\mathbf{y}}$ depends on three elements: 1. the uncertainty in the observations $\mathbf{x}$, 2. the structural uncertainty in the model $\textit{F}$, and 3. the uncertainty in the training data $\textbf{\textit{D}}$.

\subsection{Related work}
\subsubsection{Uncertainty analysis in AI \& EO}
As mentioned earlier, UQ in AI is categorized into two main types: aleatoric (or data) uncertainty and epistemic (or model) uncertainty \cite{gawlikowski2022survey}. The former captures the inherent randomness or variability in the data. It is often modeled by considering the noise or errors in the observed data. Epistemic uncertainty, on the other hand, captures the uncertainty resulting from limited/incomplete data or the assumptions made by the AI model. Typically, these two uncertainties are modeled separately using well-known UQ methods including Bayesian inference \cite{neal2011mcmc,mackay1992practical,tishby1989consistent,gal2016dropout} which assigns probability distributions to the network weights, ensemble methods \cite{lindqvist2020general} \cite{lakshminarayanan2017simple} which average predictions from multiple models, test time augmentations \cite{lyzhov2020greedy} which enable exploration of different views via augmented test samples, and single deterministic networks \cite{van2020simple,sensoy2018evidential} that capture only a single point estimate rather than a full probability distribution. 

UQ plays an important role in AI by providing insights into the reliability and robustness of the AI predictions. Without UQ, a model prediction is, effectively, uninterpretable: the uncertainty could be larger than the signal. Since these AI models often deal with complex and noisy data, their predictions are affected by uncertainties caused by various factors \cite{gawlikowski2022survey}, e.g., variability in real world situations, errors inherent to the measurement systems, incorrect training procedure, misspecification of the model architecture, or errors caused by unknown data. UQ aims to estimate and characterize these uncertainties on the predictions (also referred to as \textit{predictive uncertainty}) to make informed decisions and improve the trustworthiness of AI systems. 
Moreover, reliable estimates of uncertainty along with robust geo-variables derived from EO data may be embedded into process models (e.g., ocean, hydrological, weather, climate, etc.) to derive information vital for drawing meaningful conclusions in a wide range of applications. However, while there are numerous publications relating to AI for EO, the literature addressing UQ in EO is limited \cite{ruswurm_model_2020, bamber2022, koller2022going, hechinger2023,  ebel2023, stark2024}. 

\subsubsection{Benchmark datasets in EO}
Benchmark datasets serve as valuable resources for researchers in AI. Datasets like DeepGlobe Land Cover Classification \cite{DeepGlobe18}, SpaceNet \cite{the_spacenet_catalog_spacenet_2018}, DOTA \cite{Xia_2018_CVPR} and many more contains high quality reference labels. They enable fair comparisons and evaluations of different models and methodologies, facilitating the advancement of this field. 

One of the major hurdles in UQ for EO is the lack of dedicated benchmarking datasets. A recent publication \cite{earthnets4eo} comprehensively reviewed over 500 datasets in AI for EO, containing nearly all available resources. However, none were specifically designed for UQ. Only a handful of them mention the topic at all. To our knowledge there is no EO dataset tailored for benchmarking uncertainties of AI models. This gap exists because creating quality EO datasets with accurate labels is already a labor intensive process. The additional challenge of incorporating high quality uncertainty labels further increases the complexity and effort involved. Nevertheless, we present several existing EO datasets below. Although not specifically designed for UQ, they are still relevant and worth mentioning. 

\begin{itemize}[leftmargin=*]

    \item \textbf{So2Sat LCZ42} \cite{zhu_so2sat_2020}: So2Sat LCZ42 is the first EO dataset that provides a confidence of the labels. The authors of the dataset let a group of remote sensing experts to cast 10 independent votes on a subset of the patches in the dataset. A human confusion matrix can be obtained from these multiple label votes instead of the usual one-hot label. An overall label confidence of 85\% was achieved. These multiple labels can be fused into a distributional label that could be employed successfully for improving the model calibration and generalization performance \cite{koller2022going}. 

    \item \textbf{DroneVehicle} \cite{sun_drone-based_2022}: This dataset collects 28,439 RGB-Infrared image pairs with annotated vehicle bounding boxes for benchmarking vehicle detection algorithms. It covers various scenarios such as urban roads and residential areas from day to night. Although not containing specific label or image uncertainty information, it included an uncertainty-aware cross-modality vehicle detection framework to extract complementary information from the RGB and infared modalities.
    
    \item \textbf{INTERACTION} \cite{zhan_interaction_2019}: This dataset consists of interactive driving scenarios containing motion data of vehicles and semantic information of the maps from different traffic scenarios and different geographical locations cross the globe. The dataset does not contain uncertainty of the labels. But the authors designed a Bayesian network to provide the probabilities of the intentions of the vehicles.  





\end{itemize}

Those datasets are the early ones that take uncertainty into consideration. However, the current development is far from sufficient for a fair comparison among different UQ algorithms and their wide adaptation in the EO field.

\subsection{Contribution of this paper}
This paper presents three novel EO datasets specifically designed for benchmarking uncertainty estimates from machine learning models. These datasets address three common problems in EO: regression, segmentation, and classification. We chose biomass regression, building footprint segmentation, and local climate zones (LCZs) classification as examples to create our benchmark datasets, considering their current popularity and high relevance within the EO community. 
The design of these  datasets prioritizes aleatoric uncertainty over epistemic uncertainty. We believe that in big data regimes, modeling aleatoric uncertainty is more fundamental to real world problems themselves \cite{kendall2017uncertainties}. Aleatoric uncertainty is inherent to the data and cannot be reduced, whereas epistemic uncertainty can often be mitigated with larger amounts of data. The three proposed datasets are as follows.

\begin{itemize}[leftmargin=*]

    \item RegressionUQ: In EO, the task of biomass regression refers to estimating tree biomass from observation of tree physical dimensions. A well established method in the forestry field is use of the so called allometric equations. The RegressionUQ dataset for biomass regression simulates single tree biomass from given tree dimensions via defined allometric equations. We regard the allometric equations as the true physical model, in order to obtain the true output errors based on different input noise. Therefore, the dataset provides not only ground truth biomass, but also precise aleatoric uncertainty of the predictions at different noise level of the input data, allowing precise benchmarking of UQ methods.
    
    \item SegmentationUQ: Building segmentation takes remote sensing imagery as input and predict a binary mask of building and non-building. In a similar manner as RegressionUQ,  we construct the SegmentationUQ for building segmentation using a simulation approach. We employ very high quality building models and aerial images to render the reference image and building segmentation label in 3D modeling software. Different types of realistic noise were simulated in the input image. The corresponding aleatoric uncertainties were calculated from this. In contrast to the RegressionUQ dataset, we also simulated different label noise in the segmentation labels, which happens often in realistic training datasets. Although the dataset was created based on real aerial images and official building models, the data source is of high quality, and can be viewed as nearly noise free. This allows the analysis of the effect of different types of input and label noise (even with the same IoU) to the prediction of uncertainty by different methods. 
    
    \item ClassificationUQ: Label error is the most common noise in a classification problem. The ClassificationUQ dataset for LCZs classification contains not only one but ten labels per image patch among 10 European cities. The ten labels were created by having a group of remote sensing experts cast ten independent votes on each image patch. We developed a methods to turn the multiple votes to a distributional label. This distributional label captures the inherent uncertainty in the label, which can trigger the development of novel machine learning models that employs distribution of the label or input data, and novel UQ methods.
\end{itemize}

\section{RegressionUQ: Simulated biomass dataset for UQ in regression}
\subsection{Introduction} 
Above-ground biomass (hereafter termed biomass), defined as the dry weight of the trees in a unit forest, is an important indicator for monitoring and evaluating the forests. Recent missions, such as Biomass \cite{noauthor_biomass_nodate} and the Global Ecosystem Dynamics Investigation (GEDI) \cite{noauthor_gedi_nodate}, are tailored for producing the biomass maps on a global scale to gain more knowledge of carbon cycle on our planet Earth. But due to the high cost of acquiring ground biomass measurements and the possible imperfect biomass retrieval model resulting from the scarcity of the data, UQ is crucial. For GEDI mission, it is required that the error of 80\% of the biomass estimates shall be below 20 Mg/ha or 20\% of the estimates. 

In single-tree biomass estimation, allometric equations are accepted as the state of the art \cite{chave2014improved}. The model was obtained by fitting observed tree measurements - tree height $H$, trunk diameter at breast height $D$ (hereafter simply diameter), wood density $\rho$, and tree biomass $B$ - to a linear model in $\log$ scale $\ln B=\alpha+\beta\ln(\rho D^2H)$ 
, where $\alpha$ and $\beta$ are two parameters to fitted by the data. 
Apparently, such empirical formulas are inherently uncertain due to its over simplification and limited training data. However, biomass datasets generated from allometric equations are still scars and are regarded as the best available datasets due to the high expense of data collection for the model training and testing. Acknowledging this fact, we simplify the problem in our dataset by setting the allometric equation to be the ground truth physical model, for two reasons: 1. allometric equations are accepted in the community as the best possible solution, and 2. a ground truth biomass value could be known and the ground truth predictive uncertainty could be calculated.

\subsection{Dataset generation} 

\subsubsection{General setup}

The most widely used allometric equation for a single tree in tropical forests proposed by Chave et al. \cite{chave2014improved} was employed as the ground-truth physical model in the simulation. The equation is shown in Eqn. \ref{eqn:gt_model}, where $B$ is the biomass, $D$ the diameter, $H$ the tree height, and $\rho$ the wood density. We generate a simulated biomass uncertainty dataset with pseudo data uncertainty ground truth.
\begin{equation}
\label{eqn:gt_model}
    B=f(\rho,D,H)=0.0673\times(\rho D^2H)^{0.976}
\end{equation}
For simplicity, we fixed the wood density as $0.65$, which is a typical wood density value in the Chave dataset \cite{chave2014improved}. To create a distribution of diameter $D$ and height $H$ to resemble a realistic scenario in a forest, we fitted a library of common distributions to the diameter and height data in \cite{chave2014improved}. The best fitted distribution, which is found to be Gamma distribution, was selected. The parameters of the Gamma distributions for the diameter and height are listed in Tab. \ref{tab:gamma_D_H}. The noise free $D$ and $H$ are sampled from the Gamma distributions. The true biomass value then can be calculated from give $D$ and $H$ via Eqn. \ref{eqn:gt_model}. 

To simulate measurement noise in $D$ and $H$, we set two normal distributions with zero mean $\epsilon_D\sim \mathcal{N}(0, \sigma_D)$ and $\epsilon_H\sim \mathcal{N}(0, \sigma_H)$ respectively. In reality, the standard deviation of the measurement noise is often correlated with the absolute value of tree diameter and height. Thus, in this work, the standard deviation of diameter and height is set as linearly proportional to the mean, i.e. $\sigma_{D_i}=\alpha D_i$, and $\sigma_{H_i}=\alpha H_i$, where $\alpha$ is a noise level parameter. This is equivalent to set a signal-to-noise ratio (SNR) of $1/{\alpha^2}$ for both $D$ and $H$ respectively. Those noise were added to the noise free $D$ and $H$ to create the noisy input data.

\begin{table}[]
\caption{Parameters of fitted Gamma distributions using Chave dataset \cite{chave2014improved}.}
    \centering
    \begin{tabular}{cccc}
    \toprule
        Variable & Shape & Location & Scale \\ 
        Diameter [cm] & 0.68 & 5.00 & 30.18\\
        Height [m] & 1.92 & 1.18 & 7.75 \\
        \bottomrule
    \end{tabular}   
    \label{tab:gamma_D_H}    
\end{table}

\subsubsection{Training/testing data generation}

To generate a dataset $D$ for training a deep learning neural network model $g(\cdot)$, four steps are taken: 1) $N$ pairs of tree variables $(D_1, D_2,\ldots, D_N)$, $(H_1, H_2,\ldots, H_N)$ are generated, which is elaborated later; 2)  $N\times N$ pairs of input data are generated given the two vectors 
; 3) random noise is added to the two variables as $D'_i = D_i+\epsilon_{D_i}$, $H'_i = H_i+\epsilon_{H_i}$; 4) ground-truth biomass $B$ is calculated given $D$ and $H$ according to Eqn. \ref{eqn:gt_model}. This generates $N\times N$ pairs of data $\{(D'_i,H'_i, B_i)|i=1,2,\ldots,N\times N\}$; 5) unrealistic data points are discarded using a threshold on the biomass value, since combination of large $D$ and $H$ values are very rare. 
The test set is generated in a similar way except without adding measurement error. 

Based on the above mentioned principle, we simulated 40,000 samples in total. 80\% were set as training data, and the rest for testing.  
The variable ranges of diameter and height are set as $[5 cm, 150 cm]$ and $[1.2m, 120m]$ based on the histograms in the Chave dataset. The diameter and height samples were sampled from the Gamma distributions mentioned above. Besides, the samples whose biomass value is larger than $2236.8$ kg (90th percentile of biomass values in the Chave dataset), which is rare in real cases, are excluded from this dataset. 

\begin{figure}[h]
\centering
\includegraphics[width=0.4\textwidth]{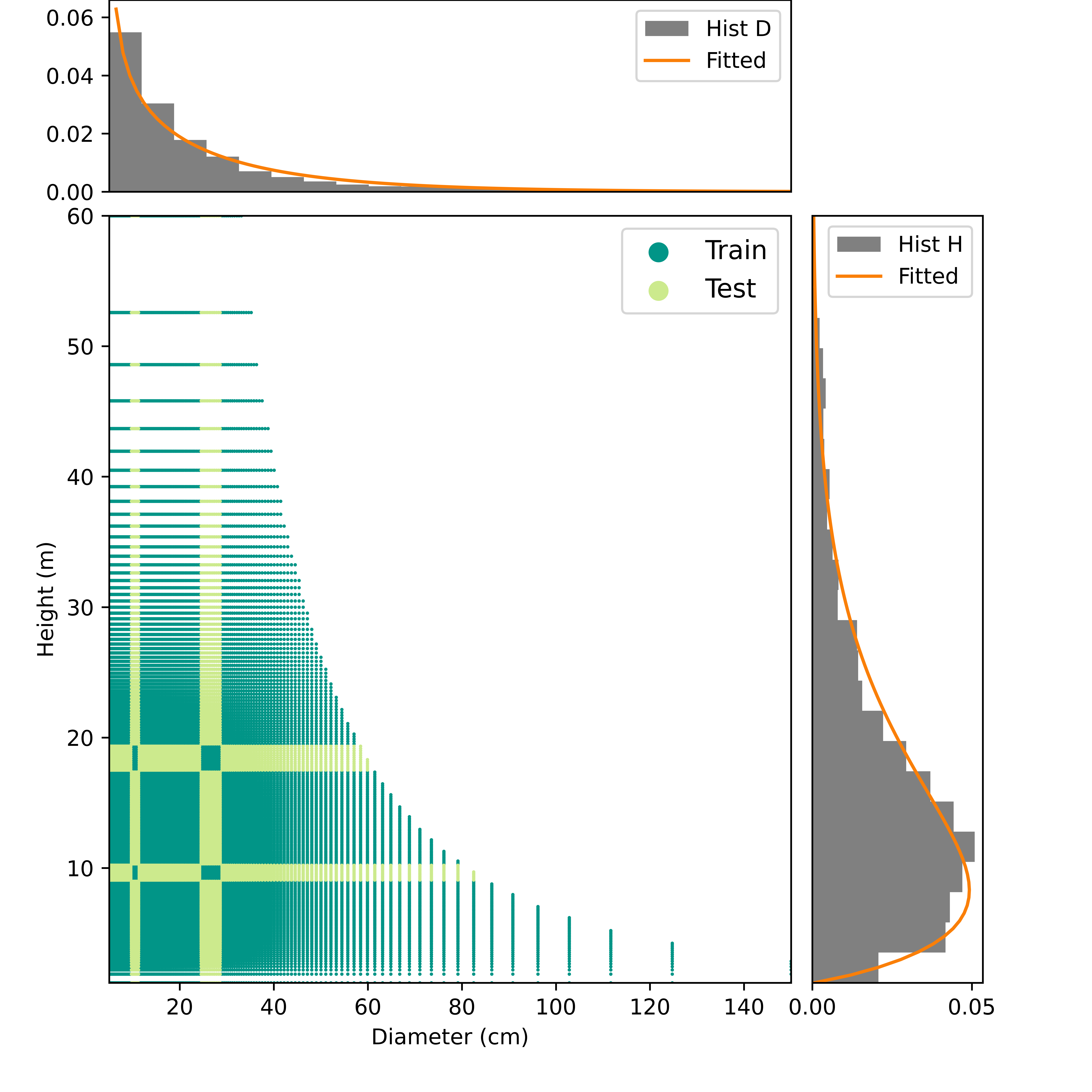}
\caption{Illustration of the $checkerboard$ training and test set split strategy on the SegmentationUQ dataset. The dots in the 2D plot are the tree diameter and height samples. They are sampled from gamma distributions fitted to the Chave dataset. Dark green and light green represent training and test set, respectively.}
\label{fig:checkerboard}
\end{figure}

A data split strategy named $Checkerboard~test$ is employed in this paper for testing the models’ predictive ability. The checkerboard test splits the input space into multiple grids. The grids are alternatively assigned as training and test set. Fig. \ref{fig:checkerboard} shows this training/test split in the SegmentationUQ dataset, where dark green is the training set, and light green is the test sets. The number of grids was set to 5 by 5. It shall neither be too small, which would make the training ans test too distinct, nor too large, which would render the distributions of training and test sets very similar.


\subsubsection{Calculation of reference aleatoric uncertainty}

\begin{figure}[h]
\centering
\includegraphics[width=0.35\textwidth]{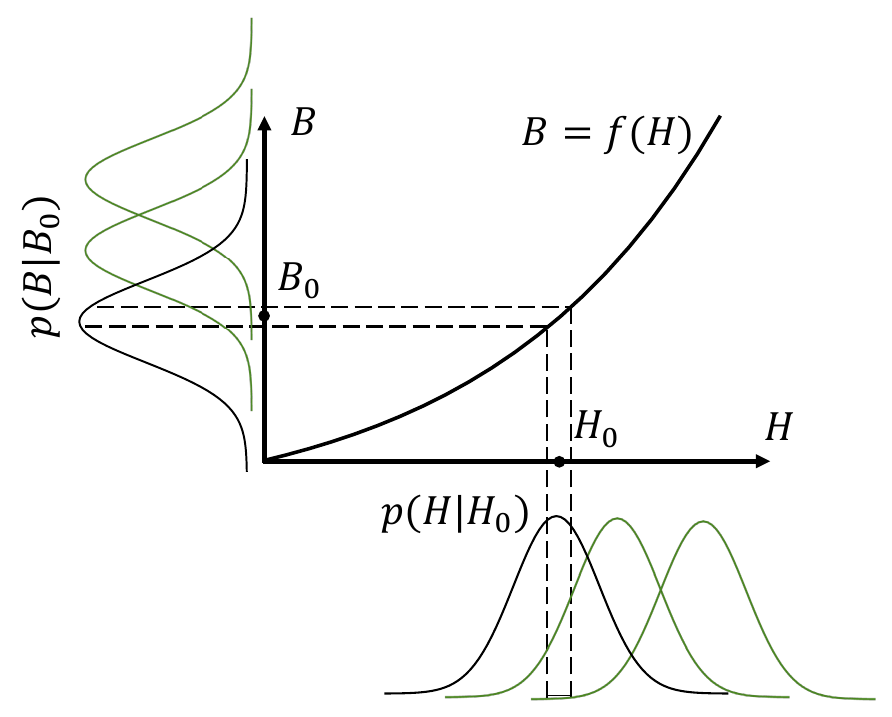}
\caption{The calculation of the reference variance of the biomass prediction. It demonstrates with a one-dimensional model $B=f(H)$. With a defined physical model, the noise at $H_0$ shown as the black Gaussian curve below the x-axis transforms to the black Gaussian curve along the y-axis through the equation. Differently, for a data-driven model trained from a wide range of data shown as a combined distribution of black and green Gaussians below the x-axis, the output uncertainty is also a mixture of all the distributions. }
\label{fig:RV_transform}
\end{figure}

With a defined ground truth physical model, it is straightforward to propagate uncertainties from the input tree height and diameter to the output biomass. Ideally, the variance on the output can be analytically calculated via random variable transformation. However, a data-driven model like a neural network is trained with the pooled training data, which is a mixture of noise distributions. Therefore, a realistic reference aleatoric uncertainty from a neural network shall also be calculated from the pooled distribution of the training data. This is explained in Fig. \ref{fig:RV_transform} using a one-dimensional model $B=f(H)$. With a clearly defined physical model, the noise at $H_0$ shown as the black Gaussian curve below the x-axis transforms to the black Gaussian curve along the y-axis through the equation. In contrast, for a data-driven model trained from a wide range of data shown as a combined distribution of black and green Gaussians below the x-axis, the output uncertainty is also a mixture of all the distributions. The realistic aleatoric uncertainty shall be estimated from the pooled output distribution. 

We employed the approach of Monte Carlo simulation by dense sampling on the input data noise and estimating the variance of the noisy output of the equation. The noise was set to be Gaussian with zero mean at each input point of tree diameter and height, and variance depending on the SNR. $20,000^2$ noisy input points were simulated for each SNR setting. The aleatoric uncertainty was estimated by calculating the variance of the equation output around its true value using 800 neighboring points. Although the estimate using 800 points is already highly accurate, a smoothing postprocessing step by fitting a parametric curve between the estimated uncertainty and inputs was followed to further increase the precision of the reference aleatoric uncertainty.

\subsection{Demonstration of the dataset}
The dataset is demonstrated by evaluating two UQ methods \cite{kendall2017uncertainties, loquercio2020general}. 
We employed a 4-layer fully-connection neural network $g(\cdot)$ as the baseline model. The network has two outputs predicting the biomass and its corresponding uncertainty denoted as $\hat{B}$ and $\hat{\sigma}_B$, respectively. The numbers of the neurons of the hidden layers are set as 16, 32, and 32 respectively. Two UQ methods \cite{kendall2017uncertainties, loquercio2020general} were evaluated in the experiment. The former one predicts the aleatoric uncertainty with an additional network output and the log-likelihood loss (Eqn. \ref{eqn:loss_gt}). The latter one \cite{loquercio2020general} uses a similar approach, but adds the variance of observations as an additional input and replaces all the layers with Assumed Density Filtering (ADF) \cite{boyen2013tractable} based layers. Because of the additional input of the observation variance, the latter UQ method is able to address the case when the noise in training and test sets belongs to different distributions. The epistemic uncertainty of both methods were estimated using Monte Carlo dropout. In essence, the two methods follow the same principle.
\begin{equation}
\label{eqn:loss_gt}
\mathcal{L}=\frac{1}{N}\sum_i\frac{1}{2}\hat{\sigma}_{B_i}^{-2}(B_i - \hat{B}_i)^2+\frac{1}{2}\log\hat{\sigma}_{B_i}^2
\end{equation}

During the training stage, 10\% of the neurons at all except the output layer are randomly dropped out. This is to prevent overfitting. At the test stage, Monte Carlo dropout was performed by sampling 90\% of the neurons of these layers $T$ times to derive multiple outputs. The variance of the outputs indicates the epistemic part of the uncertainty. The aleatoric $\hat{\sigma}_A^2$ and epistemic $\hat{\sigma}_E^2$ uncertainty were calculated as follows.
\begin{equation}
\label{eqn:ale_epi_uncertainty}
\begin{aligned}
&\hat{\sigma}_A^2 = \frac{1}{T}\sum_i^T \sigma_{B_i}^2, &\hat{\sigma}_E^2 = \frac{1}{T}\sum_i^T \hat{B}_i^2 - (\frac{1}{T}\sum_j^T \hat{B}_j)^2.
\end{aligned}
\end{equation}

In the evaluation, the prediction of the biomass and the aleatoric uncertainty were compared with our reference. R-squared score $R^2$ and relative root mean square error ($\%$RMSE) were employed as the metrics \cite{song2023biomass}. Table \ref{tab:bio_results} lists the comparison. The $R^2$ scores of biomass estimation are relatively high at all noise levels for both methods. This is mainly due to the simplicity of the task. The performance of biomass estimation from both methods naturally drops as the noise level increases, as higher noise levels in the training data leads to a noisier model. However, the predictions of aleatoric uncertainty present a different behavior than the biomass estimation. The relative RMSE values exceed $100\%$ at low noise level. The performance improves when the noise level increases to 0.10 for both methods. The performance degrades as the noise further increases beyond this value. The relatively poor performance at low and high noise levels was also mentioned in \cite{kendall2017uncertainties}, where the authors argue that the design of the loss function does not favor low and high noise levels.

\begin{table*}[]
\caption{Comparison of biomass estimation and UQ results using training sets with different SNRs and different UQ methods.\\}
\setlength{\tabcolsep}{12pt}
\centering
\renewcommand{\arraystretch}{1.5}
\label{table_results}
\begin{tabular}{cllllllll}
\toprule
   &   \multicolumn{4}{c}{ Kendall and Gal 2017 \cite{kendall2017uncertainties}}   &  \multicolumn{4}{c}{Loquercio et al. 2020 \cite{loquercio2020general}} \\
   
Noise Level   &   \multicolumn{2}{c}{$\hat{B}$}    &  \multicolumn{2}{c}{$\hat{\sigma}_B$}   &   \multicolumn{2}{c}{$\hat{B}$}    &  \multicolumn{2}{c}{$\hat{\sigma}_B$} \\
\cmidrule(lr){2-5} \cmidrule(lr){6-9}
  & $R^2$  & $\%$RMSE  & $R^2$  & $\%$RMSE  & $R^2$  & $\%$RMSE  & $R^2$  & $\%$RMSE \\
0.01    &  0.9982  & 05.76\% & -55.2789  & 999.29\% & 0.9963  & 08.27\%  & -47.8880 &  931.36\% \\
0.05    &  0.9967  & 07.76\% & 0.0212  & 133.87\%  & 0.9933  & 11.11\%  & -0.0968  &  141.71\% \\
0.10    &  0.9893  & 14.01\% & 0.9705  & \textbf{23.30\%}  & 0.9870  & 15.43\%  & 0.9776  &  \textbf{20.32\%} \\
0.15    &  0.9673  & 24.49\% & 0.7801  & 63.48\%  & 0.9795  & 19.42\%  & 0.8742  &  48.02\% \\
0.20    &  0.9556  & 28.56\% & 0.3919  & 96.41\%  & 0.9489  & 30.64\%  & 0.4940  &  87.94\% \\

\bottomrule                          
\end{tabular}
\label{tab:bio_results} 
\end{table*}

\begin{figure}[!t]
\centering
\subfloat[ Kendall and Gal 2017 \cite{kendall2017uncertainties}]{\includegraphics[height=0.20\textwidth,trim={0.6cm 0 1.6cm 0.9cm},clip]{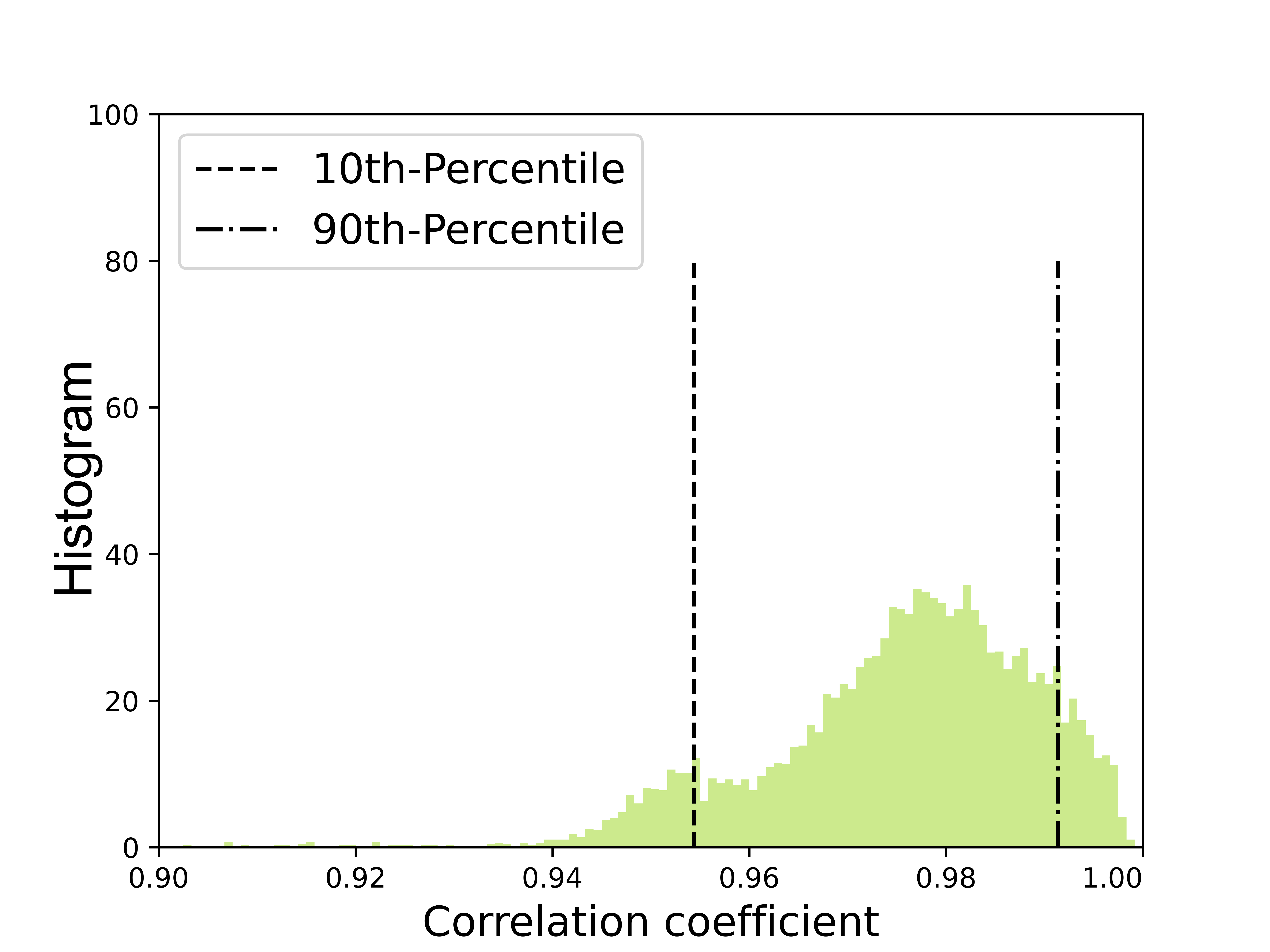}
}
\subfloat[Loquercio et al. 2020 \cite{loquercio2020general}]{\includegraphics[height=0.20\textwidth,trim={1.9cm 0 0.2cm 0.9cm},clip]{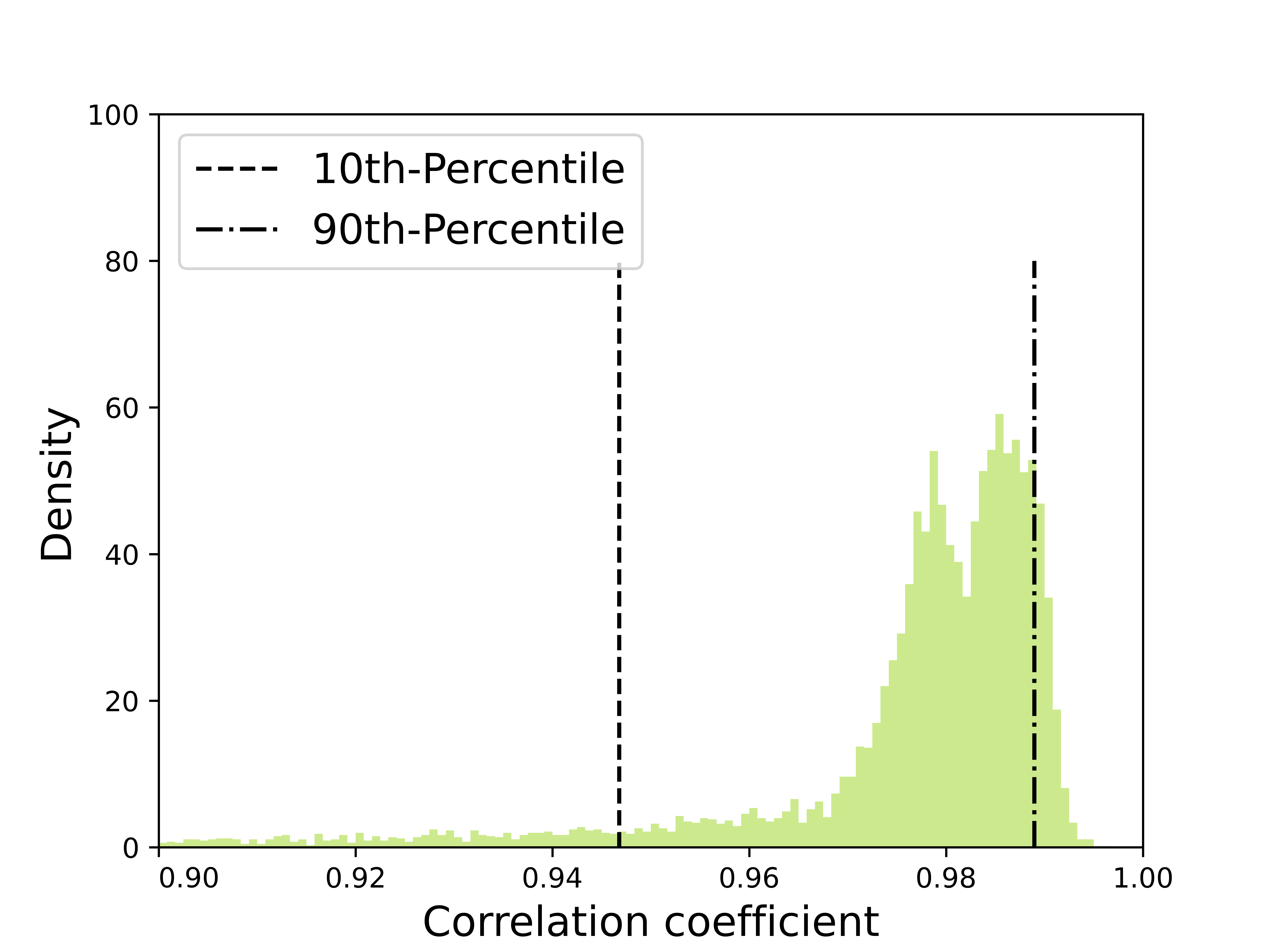}
}
\caption{Histograms of correlation coefficients between estimated aleatoric uncertainty and GT data uncertainty at pixel level using different UQ methods.}
\label{fig_bio_corr}
\end{figure}

It is evident from the results in Tab. \ref{tab:bio_results} that the latter UQ method outperforms the former. However, the \%RMSE and $R^2$ only measure the discrepancy with respect to the reference. The relative performance among the uncertainty predictions at different noise levels is also another important indicator. In order to evaluate this, we calculated the correlation coefficient between the uncertainty estimates at all noise levels and the reference uncertainty. The histogram of the correlation coefficients for the test set is shown in Fig. \ref{fig_bio_corr}. It is clearly shown that the predicted uncertainty is positively correlated to the reference data uncertainty ($90\%$ of the correlation coefficients are larger than $0.954$ and $0.947$ for the two UQ methods respectively). But the correlation coefficients of the latter method have a longer tail distribution at low values, indicating its worse performance compared to the former method. This contradictory conclusion from the performance in $R^2$ and \%RMSE is because of the relatively high bias, but more linear correlation with respect to the noise level in the results from the first method.

Thus, we demonstrate with those benchmark uncertainty values, that we can evaluate the pros and cons of different methods, and design improvements accordingly. For example, \cite{kendall2017uncertainties} has a larger bias, relative to \cite{loquercio2020general}, which could be mitigated by designing a bias correction step in the network, or via post-processing. Another deficiency of both methods is the relatively poor performance at low and high noise levels. Besides the reason mentioned in \cite{kendall2017uncertainties}, we believe it is also due to the conflation of aleatoric and epistemic uncertainty. Increasing the training size can mitigate the epistemic uncertainty, hence providing better aleatoric estimation. To explore this, we increased the numbers of both tree diameter and height samples by $N_s$ times, thus $N_s\times N_s$ times for the training data. Several models were retrained at different $N_s$ values, in order to compare the performance. The noise level was fixed at $0.10$. The same test set was used to evaluate the retrained models. The results at different training data sizes are shown in Tab. \ref{table_bio_multi}. With increasing size of the training data, both UQ methods boost the $R^2$ scores of aleatoric UQ from $0.9705$ and $0.9776$ to $0.9764$ and $0.9796$, and decreased the relative RMSE from $23.30\%$ and $20.32\%$ to $20.85\%$ and $19.36\%$, respectively. The biomass estimation results also improved. This experiment suggests that epistemic and aleatoric uncertainties can barely be disentangled in those two methods, and reducing epistemic uncertainty can potentially improve the prediction of aleatoric uncertainty. The experiments also suggest that far more training data is required to improve the prediction of the uncertainty than the prediction of the signal.

\begin{table*}[]
\caption{Comparison of biomass estimation and UQ results using different amounts of training data.}
\setlength{\tabcolsep}{12pt}
\centering
\renewcommand{\arraystretch}{1.5}
\label{table_bio_multi}
\begin{tabular}{cllllllll}
\toprule
   &   \multicolumn{4}{c}{ Kendall and Gal 2017 \cite{kendall2017uncertainties}}   &  \multicolumn{4}{c}{Loquercio et al. 2020 \cite{loquercio2020general}} \\
Training Size  &   \multicolumn{2}{c}{$\hat{B}$}    &  \multicolumn{2}{c}{$\hat{\sigma}_B$}   &   \multicolumn{2}{c}{$\hat{B}$}    &  \multicolumn{2}{c}{$\hat{\sigma}_B$} \\
\cmidrule(lr){2-5} \cmidrule(lr){6-9}
& $R^2$  & $\%$RMSE  & $R^2$  & $\%$RMSE  & $R^2$  & $\%$RMSE  & $R^2$  & $\%$RMSE \\
$\times1$    &  0.9893  & 14.01\% & 0.9705  & 23.30\%  & 0.9870  & 15.43\%  & 0.9776  &  20.32\% \\
$\times4$    &  0.9899  & 13.61\% & 0.9733 & 22.17\%  & 0.9906  & 13.14\%  & 0.9736 & 22.03\% \\
$\times16$    &  0.9941  & 10.39\% & 0.9764 & 20.85\%  & 0.9913  & 12.65\%  & 0.9796 & 19.36\% \\
\bottomrule                          
\end{tabular}
\end{table*}

\section{SegmentationUQ: Rendered dataset for UQ in image segmentation}
\subsection{Introduction}

We selected building segmentation as the example application in the segmentation dataset, as it is a common task in EO. Building segmentation is the pixel-wise classification of imagery to delineate building footprints from the surrounding environment, which holds significant importance in various applications. Several benchmark datasets have been introduced for this task \cite{earthnets4eo}. These datasets are often corrupted by noise, i.e., the real-world nature of most of these datasets increases their propensity to introduce inherent image noise, such as sensor noise or quantization error. Moreover, as the annotations are often derived from manual processes or crowd-sourced techniques, the risk of label noise being included in the dataset increases.

Despite several advancements in deep architectures of image segmentation, the sensor or label noise will always be propagated to the network's output. This introduces uncertainty in the prediction, compromising the reliability of the semantic segmentation models. Similar to the regression task, the community calls for an effective quantitative characterization of those uncertainties from reliable UQ methods. However, previous datasets lack the ground truth uncertainty vital to benchmarking existing UQ methods. Here we introduce a novel synthetic dataset, following the similar idea in the RegressionUQ dataset of employing Monte Carlo simulation of input noise and propagate to the model output. 

Since noise in aerial images can be due to various reasons, such as thermal noise and camera positioning error, 
we employed Blender, a comprehensive open-source 3D modeling and rendering software, high-quality 3D mesh models, and aerial images to simulate images with different conditions, such as varying camera viewing angles, illumination, and noise. These variations provide us with a set of noisy input images that can be used to calculate a reference uncertainty. We made use of the 3D mesh models and LoD2 building models of Berlin, Germany to render the synthetic aerial images to the corresponding 2D building masks. 10,000 image patches of different areas were rendered from the baseline ``noise-free" setting. For each of the 10,000 patches, we simulated in total 3 types of noise, 4 different noise level, with 50 random samples for each noise configuration. This amounts to total 6 million different image patches in the whole dataset.


Since the segmentation network outputs categorical predictions, it is not straightforward to generate a metric such as RMSE in the continuous domain. We then developed a novel strategy that enables quantitative comparison of UQ methods to the calculated reference uncertainty. 
\subsection{Dataset Generation}
We first present the generation of the ``noise-free" baseline dataset. Subsequently, we introduce variations of the dataset created by adding noise of different distributions into the baseline dataset, in order to investigate the effectiveness of UQ methods in quantifying noise of different distributions. In particular, we elucidate the process of calculating the reference aleatoric uncertainty for benchmarking different UQ methods.  

\subsubsection{Baseline dataset}
We generated a synthetic dataset consisting of simulated 2D aerial images and building segmentation masks. The images were rendered from 3D mesh models of Berlin, Germany using the software Blender. This allows us to simulate images with different conditions, such as camera viewing angle and position, sun illumination, and noise. The building masks were extracted from precise LoD-2 building models of the same area. The 3D models were acquired from Business Location Center Berlin download portal \cite{berlin_business_location_center}. 

\begin{table}[ht]
\caption{Baseline settings for the simulation of the synthetic dataset in Blender.}
\label{tab:blender_settings}
\centering
\begin{tabular}{l l}
\hline
\textbf{Parameter} & \textbf{Setting} \\
\hline
Camera Height & 150 m \\
Camera Angle & 90\textdegree \\
Camera Sensor Width & 36 mm \\
Output Resolution & 8192 x 8192 pixels \\
Rendering Engine & Cycles \\
Adaptive Sampling Noise Threshold & 0.01 \\
Maximum Samples & 2048 \\
\hline
\end{tabular}
\end{table}

The rendering settings for the baseline dataset are set to emulate real-world environmental conditions closely. These settings are summarized in Tab. \ref{tab:blender_settings}. This approach ensures the applicability of the dataset in real-world scenarios. It also allows us to directly vary certain environmental characteristics of the data by adjusting the parameters, thus aiding in evaluating different UQ methods across diverse environments.

The generated dataset covers a total area of 460 km² with a spatial resolution of 0.3 m. The images reflect various architectural structures, encapsulating the diverse urban landscapes within the city's terrain. Along with each image, the dataset provides a binary mask of identical size, serving as ground truth building segmentation labels. 

\begin{figure*}[t]
    \centering
    \begin{tabular}{m{0.065\textwidth} m{0.75\textwidth}}
        \small 3D VPV & \includegraphics[width=\linewidth]{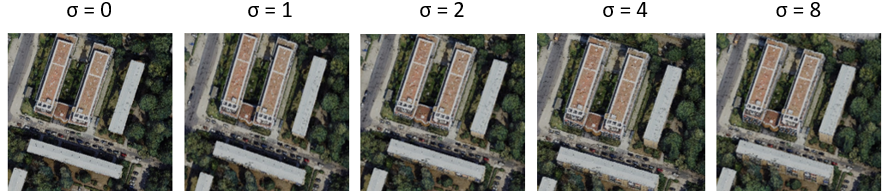} \\[-0.6ex]
        \small Gaussian & \includegraphics[width=\linewidth]{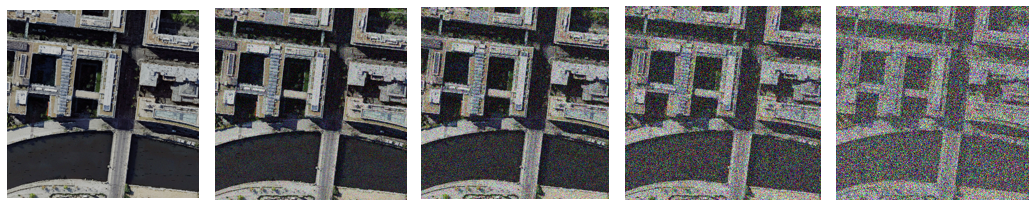} \\[-1.2ex]
        \small Poisson & \includegraphics[width=\linewidth]{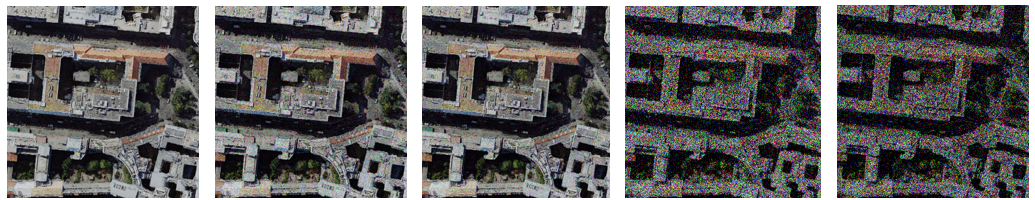} \\
    \end{tabular}
    \caption{A collection of sample variations of the dataset under three different noise types i.e. 3D viewpoint variation (3D VPV), Gaussian, and Poisson noise. Each sample is subjected to specific intensities of noise distribution determined through parameters such as standard deviation for Gaussian and lambda for Poisson distribution, set at incremental levels of 1, 2, 4, and 8. This arrangement effectively showcases the distinct impacts of image noise on the dataset, highlighting the variations in imagery under different noise conditions.}
\end{figure*}

\subsubsection{Dataset variations}




To create a dataset with different aleatoric uncertainty, we simulated variants by adding two types of typical image noise, Gaussian and Poisson, as well as by changing the camera viewing angle. These image noise types are typical image uncertainties in segmentation problems in EO. For both categories, 4 different levels of noise were simulated.

\begin{itemize}[leftmargin=*]
    \item \textbf{Image noise}: To simulate realistic image noise, we corrupted the original images with Gaussian and Poisson distributions. Gaussian noise mimics random variations in pixel values, often caused by electronic sensor noise or transmission errors \cite{corner2003noise}. Poisson noise is typically observed in images captured under low-light conditions, where the number of photons arriving at each pixel typically follows a Poisson distribution \cite{fiete2002image}. Since the pixel value of 1-byte precision ranges from 0 to 255, we applied the following clipping function after adding the Gaussian or Poisson noise.

\begin{equation}
\label{eqn:image_noise_eq}
I_{\text{Noisy}} = \max\left(0, \min\left(I_{\text{Original}} + \epsilon, 255\right)\right)
\end{equation}

In Eqn. \ref{eqn:image_noise_eq}, $I_{\text{Noisy}}$ and $I_{\text{Original}}$ are the noisy and noise-free pixel value. $\epsilon$ is the noise either in Gaussian distribution $\mathcal{N}(0, (255 \cdot n)^2)$ or Poisson distribution $\mathcal{P}(255 \cdot n)$. $n$ acts as a parameter to control the noise level, as it changes the variance of the Gaussian and Poisson distributions. 
    \item \textbf{3D viewpoint variation}: In addition to the image noise variations, we also introduced variations in our synthetic dataset by changing 3D viewpoints (or perspectives) influenced by the camera angle. This is also a typical type of uncertainty in aerial and spaceborne images, due to the errors in attitude and orbit control of the imaging platform. The camera angle directly affects the orientation of the `Camera' object in Blender used to compose shots relative to our virtual scene. A change in camera angle can therefore significantly impact the appearance of objects in the scene, such as partially obstructed buildings, less visible roof areas, and more visible walls, or objects appearing elongated or compressed. 
    
    In the simulation, we set the camera angle as a random variable sampled from Gaussian distributions. Different standard deviation of 1, 2, 4, and 8 degrees of the Gaussian distribution were simulated in order to have different levels data uncertainty. For each camera view angle variation, we rendered 50 samples. In summary, 200 variations for each scene besides the 0 degree baseline setting were rendered. 
\end{itemize}

\subsubsection{Reference uncertainty calculation}

We employ a similar MC sampling approach to the regression problem to generate the output noise distribution. One important requirement to rule out epistemic uncertainty is a ground truth physical model, which is not fulfilled for a data-driven model like a neural network. Therefore, in our implementation, we trained a baseline U-Net model $\mathcal{M}_c$ using large amounts of clean \textit{``noise-free"} data. The learning rate and batch size was set to 0.001 and 16, respectively. We believe the epistemic uncertainty of such model is minimized, as the model is well trained and fitted to the specific training data.

For the regression problem, it is obvious to take the second order moment (variance) of the prediction as the uncertainty measure, as the prediction was modeled as a Gaussian random variable.  An image segmentation network, however, outputs binary categorical values. Measuring uncertainty on a binary random variable requires modeling it as a Bernoulli distribution, which could result in loss of important information of the predicted softmax probability. Therefore, we choose the comparison to be on the distribution of the softmax probabilities. Ideally, one shall compare the predicted distribution of the softmax probabilities with a reference distribution using a metric such as the Kullback-Leibler (KL) divergence. In our binary segmentation case, the distribution can be modeled as Beta distribution, and estimated from multiple MC outputs. 
But instead of estimating a reference Beta distribution, we took a non-parametric approach of uncertainty measure. Inspired by \cite{valdenegro2022deeper}, we use the Shannon entropy (Eqn. \ref{eqn:ae}) of the distribution of the softmax probabilities as the uncertainty measure. We refer to this as the aleatoric entropy hereafter. This is a measure of the randomness in the distribution of the softmax probability. It is defined on a continuous domain, so that the predicted aleatoric entropy can be compared with the reference. 

\begin{equation}
\label{eqn:ae}
H = -\sum_{x \in \mathcal{X}} p(x) \log p(x).
\end{equation}

In order to estimate the reference aleatoric entropy given an input data distribution, we adapted the technique proposed in \cite{valdenegro2022deeper}, which represents the $j$th logit $z_j$ before the softmax layer as a Gaussian distribution $z_j \sim \mathcal{N}(\mu_j, \sigma^2_j)$. The Gaussian representation encapsulates uncertainties that can arise from sources like noisy training data or ambiguous input features. The moments of the Gaussian distribution can be estimated by feeding a set of simulated noisy input data to a ``ground-truth" model, and estimating the distribution of the logits layer. For example, for images with 8° camera view variation, we feed the 50 simulated samples to the network, and estimate the mean and variance of each logits. This probabilistic representation was employed to resample the logit many times, denoted as $z_j^1, z_j^2, ..., z_j^N$, for calculating the entropy. In our methodology, we extract a sufficient number of Monte Carlo samples, ensuring a wide coverage of the logits' distribution. Each of these samples pass through the standard softmax function, resulting in a set of probability distributions, defined by:


\begin{equation}
p_j^i = \text{softmax}(z_j^i) \quad i \in [1, N], 
\label{eqn:softmax}
\end{equation}
where $N$ is the number of Monte Carlo Samples. Since we have a binary segmentation problem, $j$ is only from 1 to 2. The second element of the softmax probabilities is always 1 minus the first element. The entropy were calculated on each logit with the $N$ probabilities. For a image patch, we obtain a single aleatoric entropy value by averaging the entropies of all the pixels and the two classes. The algorithm is illustrated in Tab. \ref{tab:ae_flowchart}.

\begin{table}[!htbp]
\caption{Estimating reference aleatoric uncertainty}
\label{tab:ae_flowchart}
\small  

\centering

\begin{tabular}{l}
\toprule
\textbf{Input:} 50 noisy patches simulated based on defined distribution\\
~~~~1. Feed the patches through the baseline model \\
~~~~2. Obtain logits tensor $\mathbf{Z}\in\mathbb{R}^{512*512*2*50}$\\
~~~~\textbf{For} each pixel \\
~~~~~~3. Estimate mean and variance of the two logits from the 50 \\
~~~~~~~~~samples. Obtain $[\mu_1,\mu_2]$, $[\sigma_1,\sigma_2]$\\
~~~~~~4. MC sample the Gaussian distribution based on the mean \\  
~~~~~~~~~and variance estimates $N$ times. \\
~~~~~~5. Calculate softmax probabilities using Eqn. \ref{eqn:softmax}. Obtain\\
~~~~~~~~~softmax probabilities $\mathbf{p}\in\mathbb{R}^{2*50}: [0,1]$\\
~~~~~~6. Calculate entropies of the two logits from the $N$ softmax  \\
~~~~~~~~~probabilities. Obtain $[H_1,H_2]$ \\
~~~~\textbf{End}\\
~~~~7. Average the entropies of all the pixels\\
\textbf{Output:} Single aleatoric entropy value $H$ for a patch \\
\bottomrule
\end{tabular}
\end{table}

\subsection{Comparison of UQ methods}

\begin{table*}[!htbp]
\caption{Comparison of the aleatoric entropy obtained from the two UQ methods with the reference across different noise types and levels in the SegmentationUQ dataset. } 
\label{table_uq_comparison_50_samples}
\centering
\small 
\renewcommand{\arraystretch}{1.3} 
\setlength{\tabcolsep}{4pt} 
\begin{tabular*}{\textwidth}{
  @{\extracolsep{\fill}}
  >{\centering\arraybackslash}m{2cm} 
  S[table-format=1.4]
  S[table-format=1.4]
  S[table-format=1.4]
  S[table-format=1.4]
  S[table-format=1.4]
  S[table-format=1.4]
  S[table-format=1.4]
  S[table-format=1.4]
  S[table-format=1.4]
}
\toprule
{Noise Level} & \multicolumn{3}{c}{Gaussian} & \multicolumn{3}{c}{Poisson} & \multicolumn{3}{c}{3D VPV} \\
\cmidrule(lr){2-4} \cmidrule(lr){5-7} \cmidrule(lr){8-10}
 & {BNN} & {TTA} & {Ref.} & {BNN} & {TTA} & {Ref.} & {BNN} & {TTA} & {Ref.} \\
0 & 0.0106 & 0.0632 & 0.0031 & 0.0106 & 0.0632 & 0.0031 & 0.0106 & 0.0632 & 0.0031 \\
1 & 0.285  & 0.229  & 0.225  & 0.317  & 0.235  & 0.209  & 0.013  & 0.086  & 0.005  \\
2 & 0.604  & 0.758  & 0.512  & 0.602  & 0.449  & 0.403  & 0.207  & 0.190  & 0.141  \\
4 & 1.471  & 1.398  & 0.906  & 1.464  & 1.125  & 1.064  & 0.489  & 0.525  & 0.474  \\
8 & 1.812  & 2.284  & 1.583  & 2.276  & 2.680  & 1.636  & 1.341  & 1.346  & 1.279  \\
\bottomrule
\end{tabular*}
\end{table*}

To demonstrate the usage of the reference uncertainty, we compare the predicted uncertainties from two UQ methods including Bayesian neural network (BNN) \cite{kendall2017uncertainties} and Test Time Augmentation (TTA) \cite{wang2019aleatoric} with our reference aleatoric entropy. 
In order for the BNN methods to generate a distribution of the softmax probability, we let the network outputs the variance of the logits, and go through step 4 and 5 in Tab. \ref{tab:ae_flowchart}. 
Tab. \ref{table_uq_comparison_50_samples} shows those aleatoric entropies for three different noise types and five different noise levels. We use the R-square score ($R^2$) and RMSE as the quality metrics to evaluate the general performance at all noise level. $R^2$ is used to assess the correlation of the prediction and the reference at different noise levels, whereas the RMSE indicates the magnitude of the errors between the ground truth and the estimated aleatoric uncertainty. In statistical terms, a lower RMSE implies a higher degree of goodness-of-fit, indicating the superior performance of the UQ method in terms of magnitude alignment with the ground truth uncertainty. Tab. \ref{table7} provides these metrics for both the UQ methods against the reference uncertainty estimate. 

Out of the two UQ methods, BNN outperforms TTA for all types of noise except Gaussian noise in terms of $R^2$. This suggests that BNN might be more adept at capturing the underlying patterns associated with aleatoric uncertainty across diverse noise conditions. Similarly, for the RMSE, BNN also demonstrates a better approximation to the reference compared to TTA for all three types of image noise. Hence, BNN outperforms TTA in this aspect as well as a lower RMSE is indicative of a model's superior precision. Moreover, we discovered that sample size increment also critically enhances the UQ method's accuracy in predicting the inherent stochastic variations. This finding aligns with the conclusion in the RegressionUQ dataset.

\begin{table}[!htbp]
\caption{Evaluation of two UQ methods using $R^2$ and RMSE under Gaussian, Poisson and viewpoint variation noise.}
\label{table7}
\centering
\renewcommand{\arraystretch}{1.5}

\begin{tabular}{lcccc}
\toprule
 Noise Type & \multicolumn{2}{c}{BNN \cite{kendall2017uncertainties}} & \multicolumn{2}{c}{TTA \cite{wang2019aleatoric}} \\
 \cmidrule(lr){2-3} \cmidrule(lr){4-5}
 &  {$R^2$} & {RMSE} & {$R^2$} & {RMSE}  \\
Gaussian &  0.9430 & 0.2771 &  0.9940 & 0.3994 \\
Poisson &  0.9996 & 0.3524 &  0.9411 & 0.4691 \\
3D VPV &  0.9981 & 0.0414 &  0.9994 & 0.0628 \\
\bottomrule
\end{tabular}
\end{table}

\section{ClassificationUQ: multiple label votes for UQ in scene classification}
\subsection{Introduction}
Remote sensing offers a wide range of applications regarding image-level classification. In the context of land use land cover classification, a popular classification scheme, the Local Climate Zones (LCZs) scheme, was introduced in \cite{stewart2011local}. Initially created to study urban heat islands, the scheme was quickly adapted to downstream application fields in urban planning \cite{perera2018local} or city mapping \cite{bechtel2015mapping}. The scheme consists of 17 classes, which consists of 10 urban and 7 non-urban classes. Although initial studies have focused predominantly on urban areas such as cities and small villages, global LCZ maps have been generated recently \cite{zhu2022urban} \cite{demuzere2022global}. 

We choose LCZs classification to creat the ClassificationUQ dataset, as we had experience in creating a large-scale LCZs training dataset \textit{So2Sat LCZ42} \cite{zhu2020so2sat}. The So2Sat LCZ42 dataset features labeled Sentinel-1 and Sentinel-2 image pairs from carefully selected 42 urban agglomerations plus 10 smaller areas across all the continents (except Antarctica). The dataset were created by identifying and labeling homogeneous areas as polygons in each city, from which Sentinel-1 and -2 image patches of 32 by 32 pixels corresponding to an area of 320 by 320 meters were cropped out. 

\subsection{Description of dataset}
Unlike other EO datasets, a rigorous, quantitative evaluation of the labeling quality was performed in the So2Sat LCZ42 dataset. We selected 10 European cities and let a group of remote sensing experts cast 10 independent votes on each labeled polygon, to identify possible errors and assess human labeling accuracy. The ``human confusion matrix" shows our human labels achieve 85\% confidence. In this work, we publish this evaluation dataset, and describes a method to turn the multiple human votes into a distributional label, instead of the typical one-hot label. It can not only serve as a quality measure of the labeling, but also inspire new UQ algorithm for noisy labels. 

The disagreement resulting from the labeling approach is attributed to the ambiguity of the classification scheme and the human uncertainty of the accompanying labeling process. The labeling uncertainty can be seen as part of the aleatoric uncertainty of the dataset. Although the measurement of the label uncertainty could be sharpened by including more remote sensing experts, the label uncertainty stays irreducible as it displays the ambiguity of the classes and the labeling process related to the dataset. This irreducibility motivates the differentiation of the human label uncertainty from the often discussed \textit{label noise}. With the aid of crowd sourcing experiments or label correction algorithms, the amount of label noise is meant to be lowered, whereas the human label uncertainty should be explicitly targeted and taken into account. The following section demonstrates the design of a new classification algorithm the using this dataset.

\subsection{Learning with Human Label Uncertainty}
We now briefly present the approach of \cite{koller2022going}, which sets up a simple yet effective framework for deep learning with human label uncertainty. Typically, when having multiple votes to an image, one uses majority vote, and uses one-hot encoding as the training label. This method make use of the multiple vote to create a distributional encoding, instead of one-hot, as the training label.

Typical one-hot encoding of LCZ is a vector of 17*1, with only one element being 1, and rest being 0, e.g. $\bm{y}=[1, 0, 0, ... ,0]$ for class 1. With multiple votes of a image exist, we can form the distributional label 
$\bm{y}_{\text{distr}}=\bm{Y} / \, M$, where $\bm{Y}$ is the vote counts of each of the 17 classes, and $M$ is the total count of votes. Following that, the KL divergence was proposed as a loss function in the training, since the objective to be minimized is a parametric probability function. We encode the LCZ dataset as $\{\bm{x},\bm{y}_\text{distr}\}$
where $\bm{x}$ is the input Sentinel image. The the image index is ignored here without losing generality. Given the predictive distribution of the neural network $f_{\theta}(\bm{x})$, denoted by $p_{\theta}(\bm{y}|\bm{x})$ (softmax probabilities), the training loss 
reads as follows: 

\begin{equation}
\label{eqn:LCZ_loss}
\mathcal{L}_{KL} (f_{\theta},\bm{x},\bm{y_\text{distr}}) =-\displaystyle\sum_{k=1}^{17} \bm{y}_\text{distr}(k) \cdot \log \frac{\bm{y}_\text{distr}(k)}{p_{\theta}(\bm{y}=k|\bm{x})}
\end{equation}

The distributional learning approach can be directly benchmarked against the classical approach using one-hot encoded labels. These can be derived by taking the majority vote of the individual experts. Model test performance results of the two approaches, using the core network architecture from \cite{qiu2020framework}, are displayed in Tab. \ref{table:ce20} (see \cite{koller2022going} for the experimental details). The cross-entropy (CE) between the model prediction and the labels is derived both for the one-hot labels (majority vote) and the distributional label explained earlier, and can be regarded as a generalization measure. Next to accuracy measures, the expected calibration error (ECE) \cite{guo2017calibration} yields the calibration performance of the model. In short, the ECE describes the discrepancy between the model's confidence (highest softmax probability) and the corresponding accuracy.  

\begin{table*}[h!]
    \small
    \centering
    \begin{tabular}{lccccc}
    \toprule \addlinespace
    & CE One-hot $\downarrow$ & CE Distr. $\downarrow$ & ECE $\downarrow$ & OA $\uparrow$ & WAA $\uparrow$ \\ \addlinespace
    \cmidrule{2-6} 
    One-hot & 1.12 $\pm$ 0.05 & 1.38 $\pm$ 0.07 & 9.79 $\pm$ 3.18 & 68.4 $\pm$ 5.5 & 69.6 $\pm$ 2.2  \\ 
    Distr. & $1.06\pm0.07$ & $1.21\pm0.07$ & 5.80 $\pm$ 1.07 & 67.0 $\pm$ 2.2 & $71.0 \pm 0.5$ \\ 
    \bottomrule
    \end{tabular}
    \vspace{0.2cm}
    \caption{Condensed performance metrics on held-out test dataset from \cite{koller2022going}. CE = cross entropy, ECE = expected calibration error, OA = overall accuracy, WAA = weighted average accuracy. }
    \label{table:ce20}
\end{table*}

\begin{figure}[h!]
    \subfloat[One-Hot Encoding]{
    \includegraphics[height=0.32\textwidth,trim={0 0 0.5cm 0},clip]{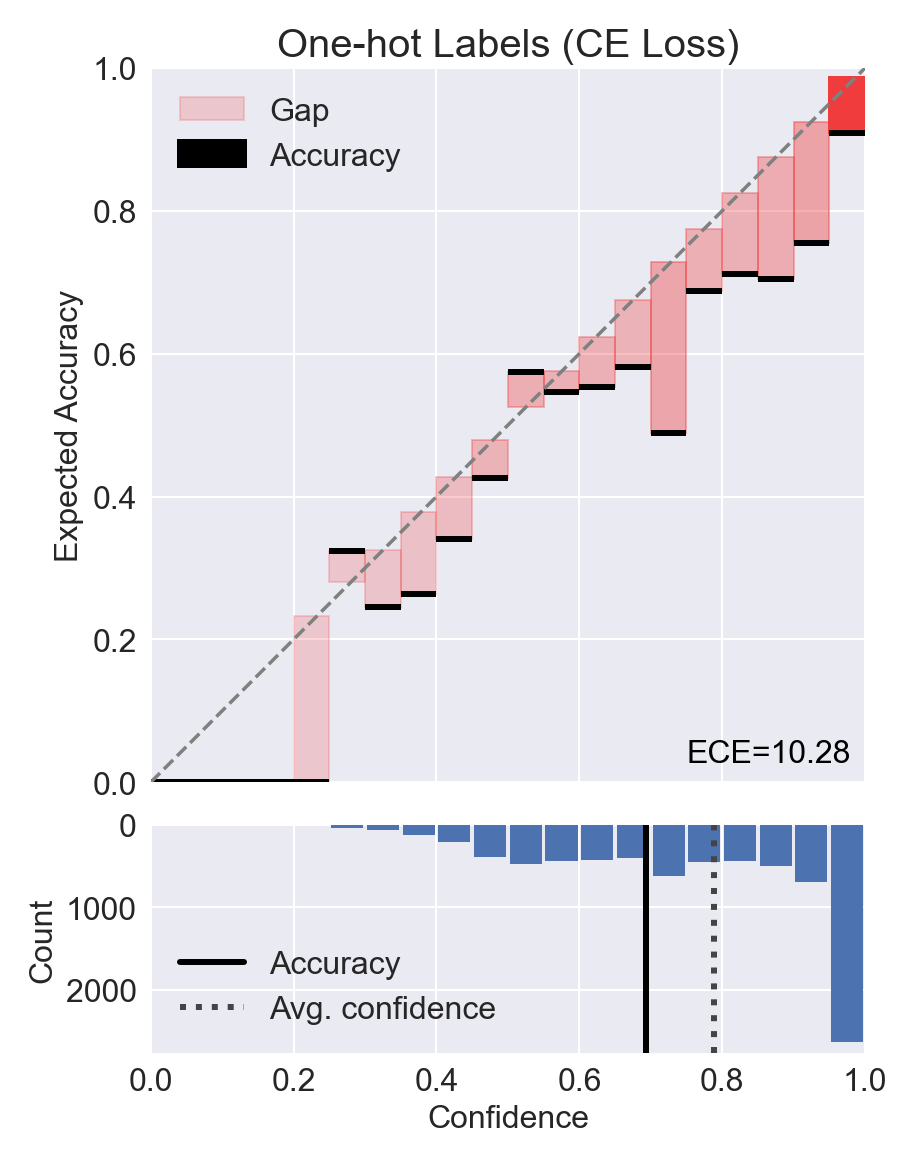}
    }
    \subfloat[Label Distribution Encoding]{
    \includegraphics[height=0.32\textwidth,trim={0 0 0.5cm 0},clip]{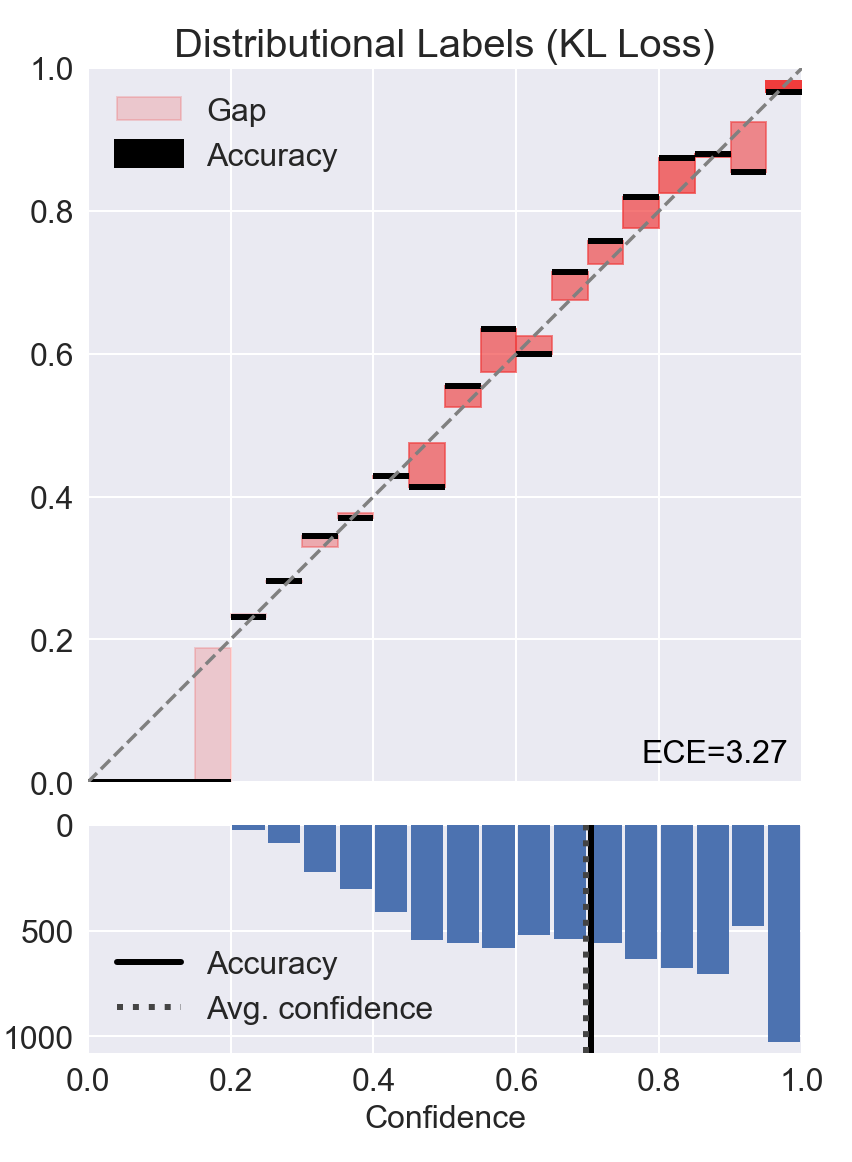}
    }
    \caption{Exemplary reliability diagrams as shown in \cite{koller2022going}.}
    \label{fig:reliability_test}
\end{figure}

The distributional learning approach has little impact on the model's accuracy, which is not surprising as the learning objective is not changed in that regard. What is striking, on the other hand, are the performance gains in terms of generalization and calibration. A more in-depth look on the model calibration is provided by Figure \ref{fig:reliability_test}. The \textit{reliability diagrams} display the binned confidences and accuracies of the two learning approaches on the test set. Every deviation from the diagonal increases the ECE, which on average is almost halved by the distributional learning approach. In conclusion, the overconfidence which is typical for convolutional neural networks can be effectively tackled by incorporating the human label uncertainty. Furthermore, generalization performance by means of lower predictive cross-entropy is optimized as well. 

\section{Conclusion and outlook}  
This article presents three datasets for benchmarking uncertainty estimates from machine learning models in EO. The three datasets covers three typical problems in EO, which are regression, segmentation and classification. Our datasets not only provide the reference labels like other EO datasets do, but also provides the reference aleatoric uncertainty of the predictions (for regression and segmentation datasets) or the label uncertainty (classification dataset). This allows users to benchmark the performance of different UQ methods.

To produce the reference aleatoric uncertainty we employ the approach of Monte Carlo simulation of the noisy input data, and propagate them through a reference model. In our work, the aleatoric uncertainty of the regression problem is measured by the variance of the prediction, whereas in the segmentation problem it is measured by the Shannon entropy of the distribution of the softmax probability. We developed a workflow to derive a reference distribution of the softmax probability given a input data distribution, based on which entropy can be computed. Although we only demonstrated on the segmentation dataset, the approach can be easily extended to a classification problem. Based on those reference uncertainties, we benchmarked popular UQ methods applied to our regression and segmentation dataset. For the classification dataset, we demonstrated a method of employing the provided label uncertainty in the training, in order to reduce the calibration error of the model.

However, there are also limitations in our approach. One needs a ground truth model through which noisy inputs can be passed, so that the aleatoric uncertainty can be isolated. This may be feasible for problems with a well understood (i.e. perfect) physical model, yet impossible to achieve for problems like classification and segmentation. An inaccurate model causes the conflation of aleatoric and epistemic uncertainties, which was observed in both the experiments for the regression and segmentation datasets. Our mitigation strategy is to employ more training samples. In terms of the classification dataset, the limitation is the high labor cost for casting multiple votes to each image, which prevents the generation of a large dataset.

\appendix[Author contribution]
Conceptualization: X.Z.; methodology: Y.W., Q.S., D.W., M.S., C.K., X.Z.; software: Y.W., Q.S., D.W., C.K.; results validation: Y.W., M.S., X.Z.; analysis: Y.W., Q.S., D.W., M.S., C.K., X.Z.; data collection: Y.W., Q.S., D.W., X.Z., the authors of \cite{zhu2020so2sat} for the human labeling in the scene classification dataset; writing, and original draft preparation: Y.W., Q.S., D.W., M.S., C.K., J.B., X.Z.; paper revision: Y.W., J.B., X.Z.; visualization: Y.W., Q.S., D.W., C.K.; funding acquisition: X.Z.; Project administration: X.Z., Y.W.; Resources: X.Z.

%



\ifCLASSOPTIONcaptionsoff
  \newpage
\fi



%
\bibliographystyle{IEEEtran}
\bibliography{ieee_template}
\end{document}